\documentclass[10pt,twocolumn,letterpaper]{article}
\usepackage{cvpr}
\usepackage{times}
\usepackage{epsfig}
\usepackage{graphicx}
\usepackage{amsmath}
\usepackage{amssymb}
\usepackage{algorithm,graphicx,xcolor,tabularx}
\usepackage{algorithmic}
\usepackage{subfigure}
\usepackage{color}
\usepackage{booktabs}
\usepackage{threeparttable}
\usepackage{enumitem}
\usepackage{times}
\usepackage{helvet}
\usepackage{courier}
\usepackage{tabularx,url,array,nth,graphicx,subfigure}
\usepackage{booktabs}
\usepackage{threeparttable}
\usepackage{enumitem}
\usepackage{color,xcolor}
\usepackage[breaklinks=true,bookmarks=false]{hyperref}
\cvprfinalcopy 
\def\cvprPaperID{1633} 

\newcommand{\wx}[1]{\textcolor{black}{#1}}
\newcommand{\czq}[1]{\textcolor{black}{#1}}

\ifcvprfinal\fi
\begin{document}

\title{On the Selection of Anchors and Targets for Video Hyperlinking}
\author{	
Zhi-Qi Cheng\textsuperscript{1,2}, Hao Zhang\textsuperscript{1}, Xiao Wu\textsuperscript{2}, Chong-Wah Ngo\textsuperscript{1}\\
\textsuperscript{1}City University of Hong Kong, \textsuperscript{2}Southwest Jiaotong University\\
{\tt\small zhiqicheng@gmail.com;
	hzhang57-c@my.cityu.edu.hk;
	 wuxiaohk@swjtu.edu.cn; 
	cscwngo@cityu.edu.hk}
}
\maketitle
\thispagestyle{empty}


\begin{abstract}
A problem not well understood in video hyperlinking is what qualifies a fragment as an anchor or target. Ideally, anchors provide good starting points for navigation, and targets supplement anchors with additional details while not distracting users with irrelevant, false and redundant information. The problem is not trivial for intertwining relationship between data characteristics and user expectation. Imagine that in a large dataset, there are clusters of fragments spreading over the feature space. The nature of each cluster can be described by its size (implying popularity) and structure (implying complexity). A principle way of hyperlinking can be carried out by picking \czq{centers} of clusters as anchors and from there \czq{reach out} to targets within or outside of clusters with consideration of \czq{neighborhood} complexity. The question is which fragments should be selected either as anchors or targets, in one way to reflect the rich content of a dataset, and meanwhile to minimize the risk of frustrating user experience. This paper provides some insights to this question from the perspective of hubness and local intrinsic dimensionality, which are two statistical properties in assessing the popularity and complexity of data space. Based these properties, two novel algorithms are proposed for low-risk automatic selection of anchors and targets.	
\end{abstract}

\section{Introduction}
Link traversal is a usual practice for web page surfing. Extending such functionality for exploratory browsing of videos remains a new problem. Video hyperlinking, a new task initiated by \cite{Trecvid2016,14_MediaEval,over2015trecvid}, aims to enhance the accessibility of large video dataset by creating links across fragments of different videos. The usage scenario is that, rather than passively searching for video content of interest, users can comfortably jump between videos by traversing hyperlinks. Generally speaking, two fundamental problems of video hyperlinking are the selection of link anchors and targets, which \czq{refer} to the sources and destination of hyperlinks. Ideally, anchors are ``hubs'' which lead users to different corners of the dataset, while targets are ``authorities'' which provide either detailed explanation of a concept or contextual information relevant to anchors.

The current research efforts on video hyperlinking include user-centric sampling of link anchors \cite{13_linkinside}, 
contextual modeling of anchors \cite{15_semantic,16cmu}, the relationship between searching and hyperlinking \cite{14_linkcontext,13_ICMR_Hyperlinking_Search} and exploitation of rich multi-modal content in video
\cite{14_linkautomatic,17_exploiting, 13_linkmultimedia}. One common theme among these studies is the query formulation of anchors for search of link targets. The employed techniques range from query expansion through linked data such as DBpedia and Freebase \cite{13_linkmultimedia}, term weighting for speech transcript \cite{14_linkautomatic} and selection of audio-visual concept classifiers for retrieval \cite{14_linkautomatic}. These efforts are mostly dedicated to the search of targets, either from the perspective of content relevancy to anchors \cite{16cmu} or link diversity \cite{17_exploiting}. To the best of our knowledge, except the ad-hoc heuristics presented in MediaEval benchmark evaluation \cite{15_SAVA}, there is no formal study on the automatic identification of link anchors from large dataset.

This paper addresses the problem of both anchor and target selections from the viewpoint of data {\em popularity} and {\em uncertainty}, an issue not yet been considered in the literature. Popularity characterizes the frequency of signals, specifically hubs and outliers in this context, of a dataset. 
\czq{Naively}, hubs could directly correspond to anchors and targets to encourage users to explore the major information sources in the close world of a dataset. Multimedia data, nevertheless, are unstructured and unordered in nature. Due to the lack of powerful techniques for semantic and context understanding, blindly bridging hubs can easily end up with \czq{the} excessive number of false and redundant linking. As the purpose is essentially about recommendation, hyperlinking should be fundamentally different from video retrieval which provides a long list of search items for user selection. In principle, recommending false or redundant link will adversely impact user experience. It is easy to imagine the frustration of users when hyperlinks bring users to ``middle of no way'' with inconsistent or repeated information.  Considering popularity without other factors such as data uncertainty and diversity could lead to meaningless hyperlinking in practice. This paper characterizes uncertainty as the complexity of {\em local} data distribution. Imagine that in a large dataset, the distribution of data changes when moving across different regions of the space. In regions of highly complex configuration, the probability of false linking will also proportionally increase. Creating hyperlinks with the potential risk in mind is thus important, for example, by introducing additional features for reducing data uncertainty. 

To this end, this paper leverages two established studies, hubness \cite{10_Hubness} and local intrinsic dimension (LID) \cite{15_LID}, to quantify anchors and targets based on statistical data properties. Hubness measures popularity by counting the number of times that a point in \czq{high-dimensional} space appears among the $k$-nearest neighbors of other points. A popular point, or hub, can carry either blessing or warning signal. In retrieval for example, certain items share high similarity to disproportionally many other items, and thus are likely to be retrieved. Proper pre-processing of these items, such as by removing them from dataset, can improve search performance as studied in music retrieval community \cite{12_Hub_mirex}. On the other hand, using hubs to initialize clustering algorithm, such as k-means, can boost performance \cite{14_Hub_role}. LID, different from global dimensionality reduction techniques such as PCA, locally quantifies the intrinsic dimensionality of a point. In theory, a point embedded in a space with high number of LID will suffer from curse of dimensionality, due to the fact that distance or similarity measure becomes less meaningful in high dimensional space. LID of a point is calculated by the rate at which the number of neighboring points grows as the range of distances expands from that point \cite{15_LID}.

By hubness and LID, this paper characterizes the statistical properties of anchors and targets in the Blip1000 dataset provided by Video Hyperlinking (LNK) of TRECVid 2016 \cite{Trecvid2016} for case study. Particularly, the inter-play among hubness, LID and diversity are explored to provide an insightful look \czq{at} the anchors, targets and multi-modal feature fusion. In addition, based on the result of analysis, we further propose two novel algorithms for automatic selection of link anchors and targets. The contribution of this paper is on quantifying the statistical properties of anchors and targets, and gives light \czq{in} the way of selecting anchors and targets based on data popularity and risk.

\section{Data Analysis}
\label{sec:Du}
\subsection{Hubness}
\label{sec:hub}
Hubness characterizes the popularity of a fragment, defined as the number of times that a fragment is regarded as the $k$ nearest neighbors of other fragments \cite{10_Hubness}. Let$ \{x_1,..., x_n\}$ be the $n$ fragments  drawn from a dataset, the hubness score of a fragment $x$ is
\begin{equation}
N_k(x) = \sum^{n}_{i=1} p_{i,k}(x)
\end{equation}
where 
\begin{align}
\label{eqn:nk}
p_{i,k}(x)=
\left\{\begin{matrix}
1, & \text{if $x$ is among the $k$ nearest neighbor of $x_i$,}
\\ 
0, & \text{otherwise}.
\end{matrix}\right.
\end{align}
Basically $N_k(x)$ counts the number of fragments that include $x$ as their $k$th nearest neighbors under a predefined distance measure. Based on \cite{10_Hubness}, the score can be utilized to categorize a fragment as hub if $N_k > k$, anti-hub if $N_k = 0$, and normal otherwise. In the context of hyperlinking, hubs are popular fragments that can potentially outreach to different ``corners'' of a dataset. A fragment with extremely high value of score, nevertheless, can be noise that might not worth linking. As studied in \cite{Hub_class}, removing these noises can improve the classification performance by 1\%-5\%.

By viewing $N_k(x)$ of all fragments as a probability distribution, the hubness of a dataset can be further characterized by the degree of skewness, defined as
\begin{equation}
\label{eqn:SNK}
S_{N_k}=\frac{E\left ( N_k-\mu_{N_k} \right )^3}{\sigma _{N_k}^{3}}
\end{equation}
which is the third moment of the distribution, with $\mu_{N_k}$ and $\sigma_{N_k}$ as the mean and standard deviation respectively. According to \cite{10_Hubness}, hubness exists in a dataset when $S_{N_k} > 1$. The bigger this value is, the less number of hubs in the dataset. In short, skewness gives a statistical sense of whether a dataset is appropriate for hyperlinking, and its degree indicates the proportion of fragments that could be regarded as hubs for hyperlinking.

\subsection{Local intrinsic dimension (LID)}
Intrinsic dimensionality refers to the minimal number of dimensions required to {\em globally} describe a dataset.  Local intrinsic dimension (LID), instead, characterizes this property {\em locally} with respect to the neighbourhood structure of a fragment $x$. The higher the value of LID is for $x$, the more difficult or risky to describe the neighbourhood of $x$ under a predefined distance function. In this paper, we employ Maximum Likelihood Estimation (MLE) \cite{15_LID} for calculation of LID. Consider $n$ number of fragments 
in the vicinity of $x$ with distance smaller than $\omega$, and let $\{l_1, l_2,..., l_n\}$ be their distances sorted in ascending order in the range of $[0,w)$. The LID of $x$ is define as
\begin{equation}
\label{eqn:IDL}
\widehat{ID_L}=-\left (\frac{1}{n}\sum_{i=1}^{n} ln\frac{l_i}{\omega } \right )^{-1}
\end{equation}
\czq{where $l_i$ is a distance which is always larger than 0. Due to the problem of floating point representation, $l_i$ may be treated as 0. In this case, $l_i$ is set to the smallest value that can be recognized by the computer.} 
LID in principle characterizes the risk of hyperlinking. Specifically, the chance of creating false links can be proportional to the number of LID. When the dimensionality is high, one may consider using a different distance measure or inclusion of multi-modal features to reduce the number of dimensions before link establishment.

\subsection{Simulation}
\label{sec:simulation}
In this section, we use hubness and LID to analyze the statistical property of the Blip10000 dataset \cite{Blip10000}, which is the \czq{same dataset used by TRECVid 2016 benchmark} for video hyperlinking \cite{Trecvid2016}. The dataset consists of 340,342 fragments from \czq{around 3000 hours} of videos.

\subsubsection{ Does hubness exist?}
As the dataset is huge, we employ sampling strategy to randomly select a subset of fragments for experiment. Three popularly employed features are considered in this simulation:
\begin{itemize}[leftmargin=*] 
	\item CNN:  Deep feature extracted from ResNet-50 \cite{Pool_5}. The pool-5 feature, which is averagely pooled from the feature maps of last convolution layer, is employed. The dimension of feature is 2,048.
	\item Concept: High-level feature composed of the classifier responses of 1,000 ImageNet concepts \cite{Con_1000}. The classifiers are \czq{learned} with ResNet-50 and the feature is in 1,000 dimensions.
	\item Text: Bag-of-words feature extracted from automatic speech recognition (ASR) transcript and video metadata. The feature is weighted with TF-IDF and is in 32,797 dimensions. 
\end{itemize}

Table \ref{tab:hub} shows the simulation result. For all the three features, the skewness values are greater than 1.0, clearly indicating the existence of hubness in the dataset. Text feature, in particular, is much skew than CNN and Concept features. The high skewness can signal the less discriminative power of text than visual features for some fragments. \wx{We pick up the fragments with high value and notice that most of them correspond to lengthy speech with general content.}
Apparently these are not ideal candidates for link anchors. 

When both visual features are early fused, the value of skewness increases. We speculate that fusion actually disambiguates some pairwise similarities due to the inclusion of new modality, resulting in less number of hubs and hence higher value of skewness. When all the three features are fused, the skewness value keeps increase. Nevertheless, the value is not as high as that of text feature alone. 
With the result, we conclude that multi-model fusion is likely to produce a better result in anchor or target selection than any of the three single modalities.

\begin{table}[!t]
	\centering
	\caption{Skewness of hub scores in Blip10000 dataset. The experiment is run for multiple times. Each time the number of fragments being sample is in the range of 11,481 to 44,602. The last column corresponds to the skewness of distribution, by setting $k=10$ nearest neighbors and using cosine similarity.
	}
	\vspace{0.1in}
	\label{tab:hub}
	\begin{tabularx}{6.5cm}{lcc}
		\hline\hline
		\centering{}& \textbf{$dim$} &\textbf{$S_{N_{10}}$}\\
		\hline
		CNN  & 2,048  & [0.94-1.83] \\
		Concept & 1,000   &  [1.42-2.03] \\
		Text & 32,797 &  [3.74-4.97] \\
		\hline
		CNN+Concept  & 3,048  & [1.42-1.96] \\
		CNN+Concept+Text  & 35,845 & [2.68-3.97] \\
		\hline
		\hline
	\end{tabularx}
\end{table}

\subsubsection {How popular and risky are the anchors?}
\begin{figure}
	\centering
	\includegraphics[scale=0.4]{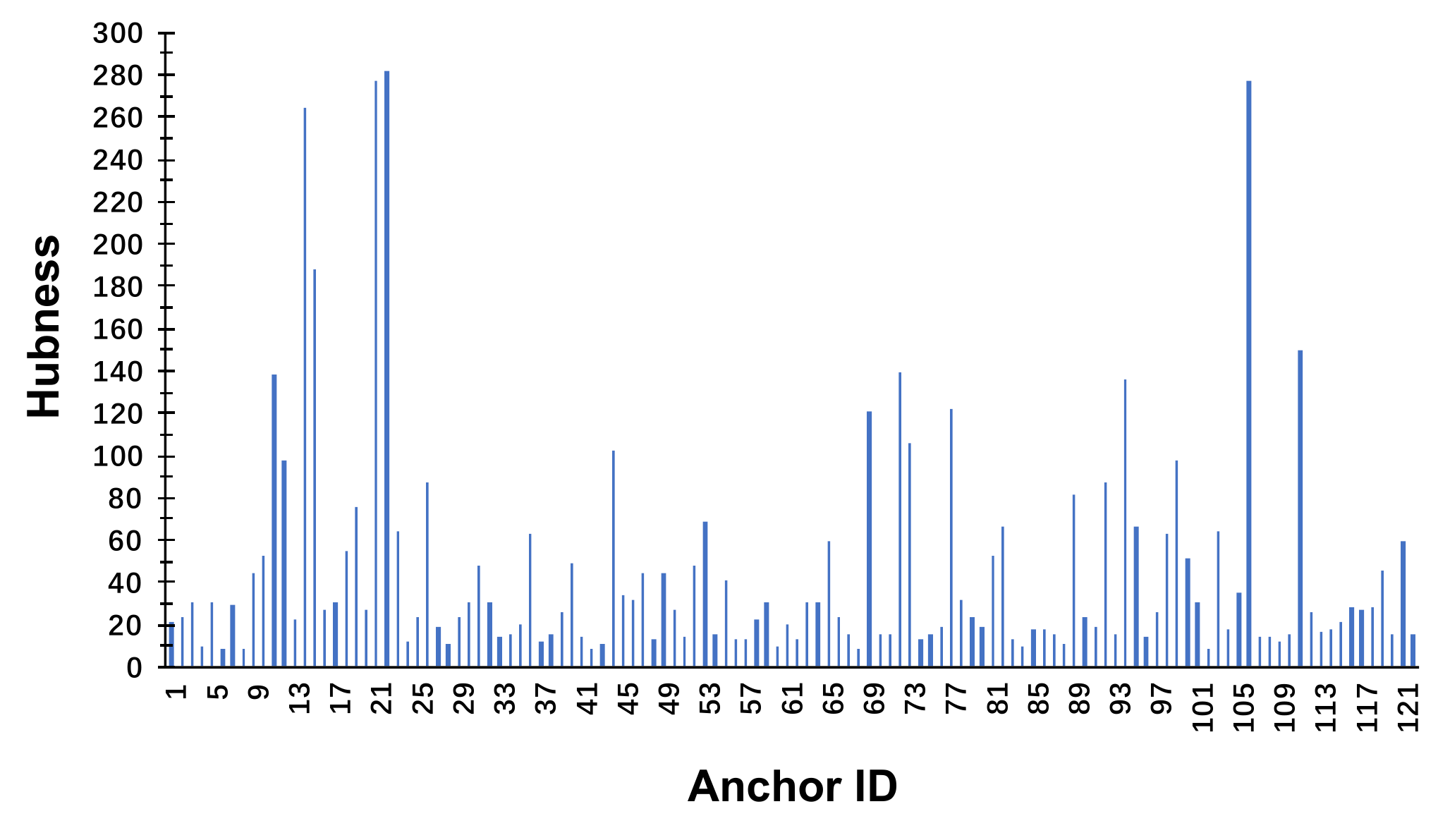}
	\caption{The hub score of 122 anchors in Blip10000.}
	\label{fig:1}
\end{figure}

We use 122 anchors provided in the dataset for simulation. These anchors are manually identified as being the fragments of interest that users would likely to further explore more information from. We calculate the hub scores of the anchors using the combination of three features over the whole Blip10000 dataset.  We set the number of nearest neighbors $k=10$, and the result is shown in Figure \ref{fig:1}. As a user is not likely to continue watching a video if no relevant content is found after few minutes, we consider only the first 3-minute of a fragment during the calculation. Among the 122 anchors, 114 of them are regarded as hubs for having values greater than 10, i.e., $N_k > k$. As shown in Figure \ref{fig:1}, majority of anchors indeed have hub score much higher than $k=10$. 

\begin{figure}
	\centering
	\includegraphics[scale=0.41]{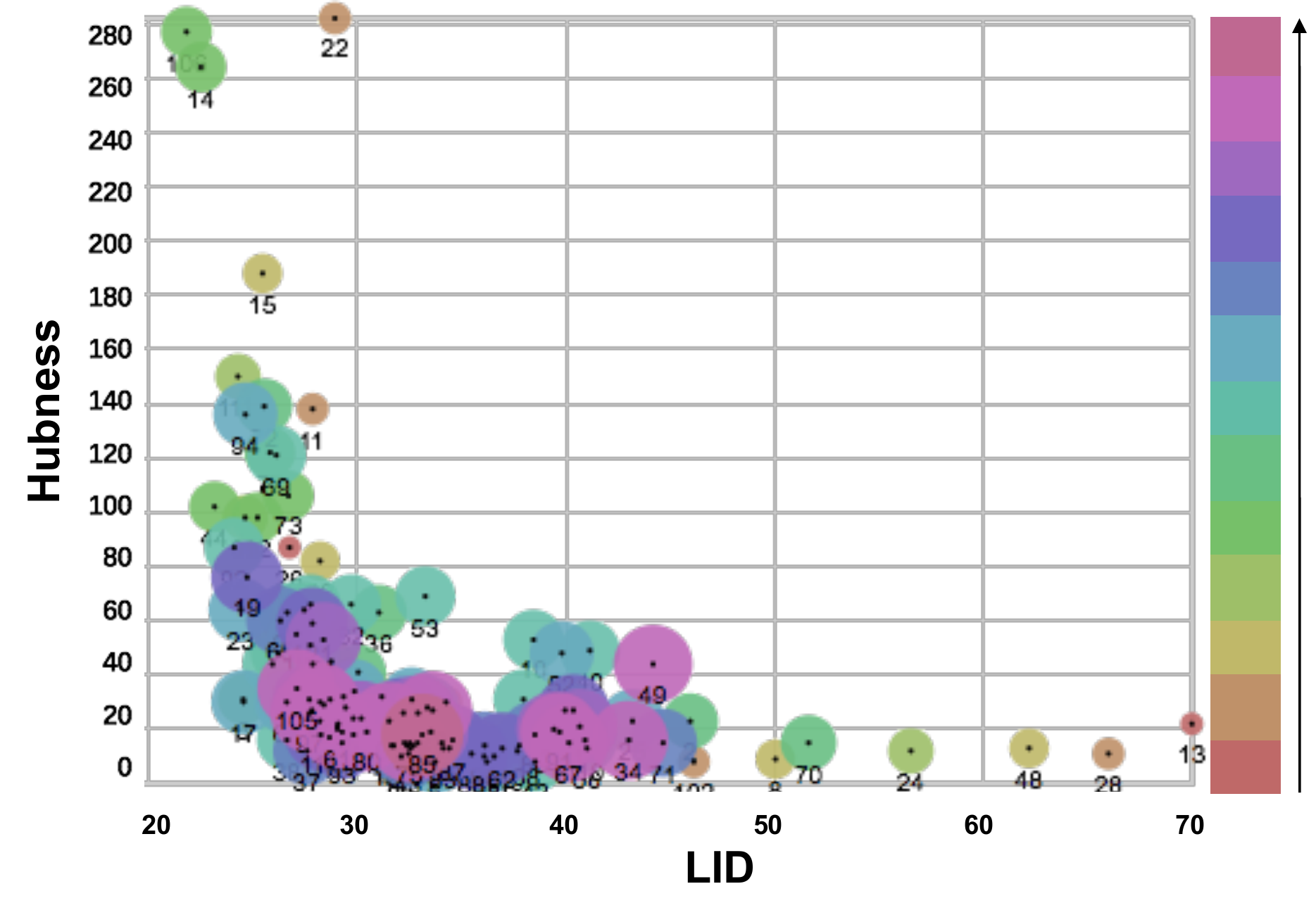}
	\caption{The relationship between hubness, LID, and diversity.
		The number indicates anchor ID.
		The color and size of radius show the diversity of an anchor.
	}
	\label{fig:2}
\end{figure}

We further investigate the interplay between hubness and LID. 
As shown in Figure \ref{fig:2}, the average LID of anchors is 33, which is significantly less than the global intrinsic dimension (53) of the whole  dataset calculated using \cite{04_ID}.
As shown in Figure \ref{fig:2}, majority of anchors are with values in the range of 20 to 40, basically showing low risk in hyperlinking in such a low dimensionality. Finally, we also take into consideration of the role of diversity in hyperlinking. We calculate diversity as the average pairwise distance between the 30 nearest neighbors of an anchor. The result is presented in Figure \ref{fig:2}, where diversity is visualized by the radius of an anchor. It is not \czq{surprising} to see that the value of diversity is somewhat inversely proportional to the hub score. Nevertheless, for anchors with higher LID value, the diversity value tends to be small. The result basically can give a clue that these are difficult anchors for hyperlinking.

\czq{
	Note that the 122 anchors actually belong to two sets. The first 28 are development anchors focusing on what people say \cite{14_MediaEval}, while the latter are 94 testing anchors conveying verbal visual information \cite{17_dataset}. We also notice some fundamental difference between the two sets of anchors. Among the anchors with hubness $>$ 122 as shown in Figure \ref{fig:1}, 58\% of them are from development anchors versus 42\% from testing. Similarly in Figure 2, for LID $>$ 50, there are 67\% (33\%) from development (testing) anchors. Based on this statistics, we speculate that testing anchors shall generate overall better performance in hyperlinking.
}

\subsubsection{How different are targets from anchors?}
\begin{figure}
	\centering
	\includegraphics[scale=0.44]{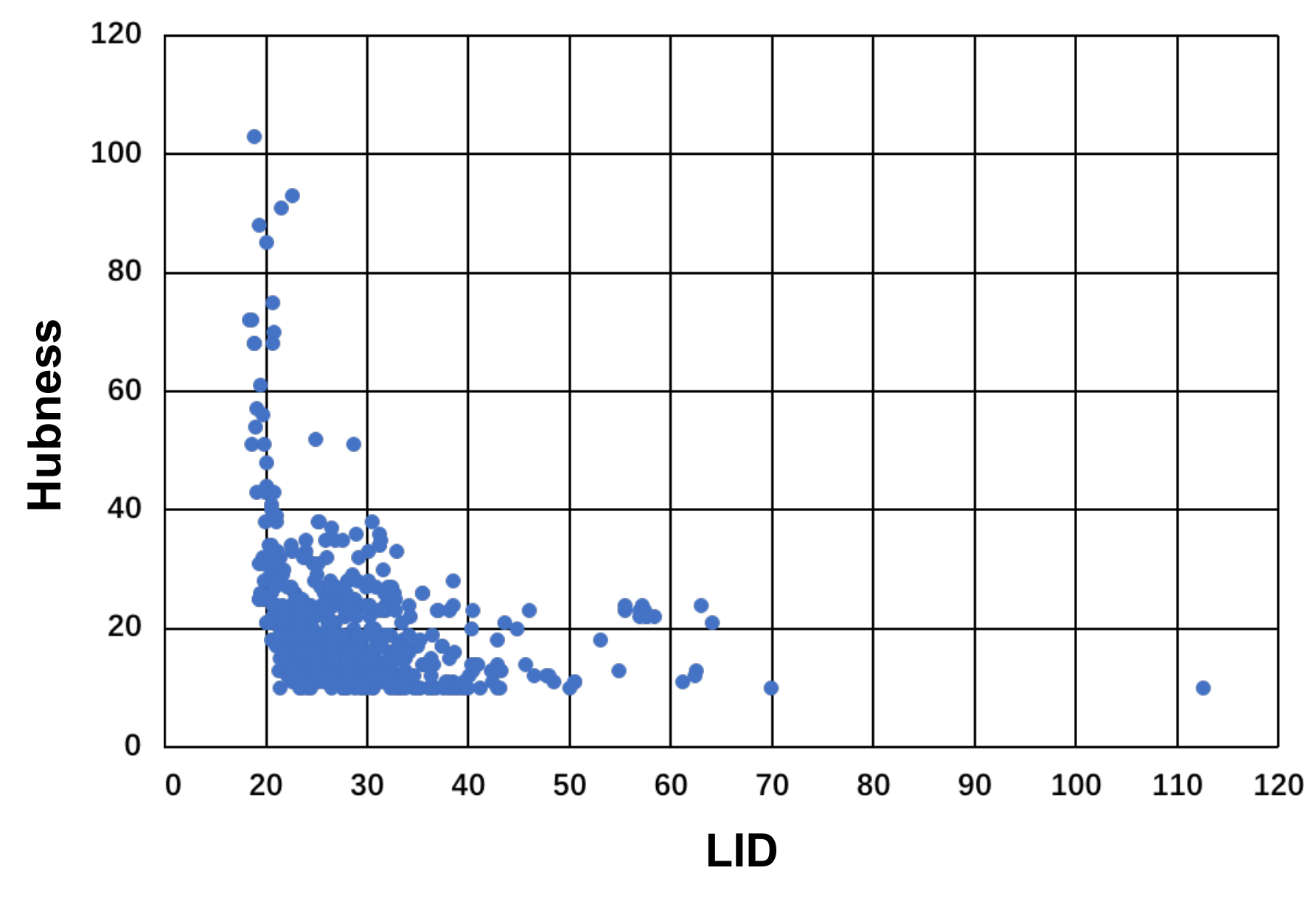}
	\caption{The distribution of 607 targets in the 2D space of LID and hubness.}
	\label{fig:4}
\end{figure}

We use 607 ground-truth link targets provided for 28 anchors in this simulation. Similar to anchors, these targets are manually judged as worth for hyperlinking. Figure \ref{fig:4} shows the distribution of target fragments in terms of hubness and LID. First, the hub scores are much lower than that of anchors. Second, a larger portion of fragments (94\%) are cluttered in the LID interval of 20-40 than anchors. An interesting observation is that, different from anchors, a lower value of LID for a target may not correspond to a higher value of hub score. This can be explained by the fact that targets are more specific in content for detailing certain aspects of anchors, and thus are with lower hub scores than anchors. Similarly as in the previous subsection, we further investigate the degree of diversity for the targets. 
The average distance among the 30 nearest neighbors of targets is larger than that of anchors. 
The result shows that the \czq{neighborhood} regions of targets are sparser or diverse.

\section{Algorithm}

We assume that the set of ground-truth anchors and targets are preferred examples for video hyperlinking. Based on the simulation results in Section \ref{sec:simulation}, we speculate that both anchors and targets should have high value of hubness and diversity, while with lower value of LID. With this, we formulate the selection of anchors and targets as an optimization problem as following.

Let $X=[x_{1},x_{2},...,x_{n}]\in R^{d\times n}$ as a dataset with $n$ fragments, where $x_i$ denotes the feature of $i$th fragment and $d$ is feature dimension. The corresponding hub scores and LID values are denoted as $H=[h_{1},h_{2},...,h_{n}]\in R^{n\times 1}$ and $D=[d_{1},d_{2},...,d_{n}]\in R^{n\times 1}$ respectively. Furthermore, let $A$ be an affinity matrix, where its element $A_{i,j}$ indicates the distance between fragments $x_i$ and $x_j$. The optimization aims to select $k$ out of $n$ fragments as either anchors or targets with the objective function

\begin{equation}
\label{eqn:obj}
\begin{split}
\underset{Y}{max}\left ( \frac{Y^{t}H}{k}- \frac{Y^{t}D}{k}  +\frac{Y^{t}AY}{k\left ( k-1 \right )} \right),
\\
s.t.\sum_i y_i=k,y_i\in \left \{ 0,1 \right \} 
\end{split}
\end{equation}
where $Y=[y_{1},y_{2},...,y_{n}]\in R^{n \times 1}$  is a binary indicator vector, with $y_i=1$ means to select fragment $x_i$ and otherwise. Basically the objective function is to sample a predefined number of fragments that are popular and diverse but with low risk in hyperlinking. 

Optimizing Equation (\ref{eqn:obj}) is difficult, nevertheless, due to the 0-1 selection constraint imposed by the indicator vector. We relax the constraint to $y \in [0,1]$, as following

\begin{equation}
\label{eqn:relax}
\begin{split}
\underset{Y}{max}\left ( \frac{Y^{t}H}{k}- \frac{Y^{t}D}{k}  +\frac{Y^{t}AY}{k\left ( k-1 \right )} \right),
\\
s.t.\sum_i y_i=k,y_i\in [ 0,1] 
\end{split}
\end{equation}
making the formulation similar to the standard quadratic programming problem. The major difference lies in the constraint $y_i \in [0,1]$ to prevent the solution from being dominated by a small number of fragments. In our solution, two Lagrangian multipliers, $\mu_i$ and $\beta_i$, are introduced for each variable $y_i \in Y$. Furthermore, an additional multiplier $\lambda $ is used for the constraint $\sum_i y_i=k$. Equation (\ref{eqn:relax}) is turned into the form of

\begin{equation}
\label{eqn:lag}
\begin{split}
\underset{Y}{max}\left ( \frac{Y^{t}H}{k}- \frac{Y^{t}D}{k}  +\frac{Y^{t}AY}{k\left ( k-1 \right )} \right)
\\
-\lambda(\sum_{i}yi-k )+\sum_{i}u_{i}y_{i}+\sum_{i}\beta_{i}(1-y_{i})
\\
s.t.\sum_i y_i=k,y_i\in [ 0,1] 
\end{split}
\end{equation}
Based on the Karush Kuhn Tucker (KKT) conditions \cite{90_KKT}, the solution that maximizes Equation (\ref{eqn:lag}) must satisfy the following first-order necessary conditions
\begin{equation}
\left\{\begin{matrix}
\left ( \frac{H}{k}- \frac{D}{k}  +\frac{2AY}{k\left ( k-1 \right )} \right)_{i}-\lambda+\mu_{i}-\beta_{i}=0
\\
\\ \sum_{i}u_{i}y_{i}=0
\\
\\ \sum_{i}\beta_{i}(1-y_{i})=0
\\
\end{matrix}\right.
\end{equation}
where the subscript $i$ can be viewed as index to a fragment. Since $y_i$, $m_i$ and $b_i$ are non-negative, the KKT conditions can be rewritten to
\begin{equation}
\label{ean:rew}
r_i(y)=\left ( \frac{H}{k}- \frac{D}{k}  +\frac{2AY}{k\left ( k-1 \right )} \right)_{i}
\left\{\begin{matrix}
\leq  \lambda & y_{i}=0
\\
\\ =  \lambda & y_{i}\in(0,1)
\\
\\ > \lambda & y_{i}=1
\end{matrix}\right.
\end{equation}
which elegantly quantifies a fragment into three categories based on the value of $\lambda$. The value returned by the function $r_i$ is treated as the ``reward'' of a fragment. To this end, we employ pairwise updating algorithm \cite{12_Dense_Neighbor} for optimization. In each iteration, two variables $y_i$ and $y_j, i \neq j$, will be updated as following
\begin{equation}
\label{eqn:med}
\hat{y_{l}}=
\left\{\begin{matrix}
y_{l} ,\; l \neq i,l \neq j;
\\
\\ y_{l}+\alpha ,\; l =i;
\\
\\ y_{l}- \alpha, \; l=j;
\end{matrix}\right.
\end{equation}
By some mathematical manipulations, it can be shown that the value of $\alpha$ is
\begin{align}
\label{eqn:para}
\alpha=
\left\{\begin{matrix}
\min(y_{j},i-y_{i}), \sigma \ge 0 \;and\; \eta>0;
\\ 
\\
\min\left (y_{j},i-y_{i}, \frac{k(k-1)(r_{j}(y)-r_{i}(y)}{A_{ii}+A_{jj}-2A_{ij}}\right),\sigma < 0 \;and\; \eta>0;
\\
\\
\min(y_{j},i-y_{i}),\sigma>0  \;and\; \eta=0;
\end{matrix}\right.
\end{align}
where $\sigma= A_{ii}+A_{jj}-2A_{ij}$  and $\eta= r_{i}(y)-r_{j}(y)$. To guarantee algorithm convergence, the pairs of fragments for updating should be carefully selected based on their rewards in Equation \ref{ean:rew}. Specifically, $r_i(y)$ should correspond to the fragment with the highest reward among the fragments whose rewards are smaller or equal to $\lambda$. While for $r_j(y)$, the fragment with the smallest reward among those whose rewards are larger or equal to $\lambda$ should be selected.


\section{Selection of anchors and targets}
A critical step that governs the selection of fragments is the initialization of the binary indicator vector $Y$. In principle, the initialization tunes the optimization to favor selection of certain fragments. In the implementation, we leverage this trick for the selection of anchors and targets. We propose two algorithms: Hub-first and LID-first, where the former (latter) prioritizes fragments higher (smaller) in hubness (LID). The Hub-first algorithm initializes $Y$ by setting $y_i = 1$ for variables that correspond to the fragments with the first $k$ largest hub scores. Similarly for the LID-first algorithm which sets $y_i = 1$ for fragments corresponds to the first $k$ smallest LID values. Based on the simulation results in \ref{sec:simulation}, anchors are \czq{preferable} to be high in hubness and low in LID. In such case, ideally either hub-first or LID-first is appropriate for identification of anchors. On the other hand, as targets are more specific in content, their hub scores are generally not as high as anchors. LID-first appears to be more appropriate for target selection.

\section{Experiments}
The experiments are conducted on \czq{the full} Blip10000 dataset \cite{Blip10000}. The hubness, LID, and diversity of each fragment are offline precomputed over the whole dataset.

\subsection{Anchor selection}
The experiment assesses the ability \czq{to} rank the 122 anchors provided by TRECVid LNK. As there is no ground-truth of whether one anchor is better than another for hyperlinking, we conduct the user study to evaluate the quality of anchors. Ideally, an anchor should deliver the message that users would like to further explore. A necessary though not sufficient condition is that such message should be unambiguous and explicit to most users. 
\czq{In this study, we recruit a total of 15 graduate students as evaluators, who are not native English speakers. They are instructed to watch 122 anchor fragments and then note down the messages of anchors in words. Note that the students are not familiar with video hyperlinking and Blip10000 dataset. Hence, the messages represent the understanding of key video content from their own perspectives.}
For each anchor, we manually check the 15 descriptions and then assign a score in the range between 0 and 15, indicating the number of descriptions that are consistent in meaning.  Through this process, the quality of each anchor is reflected by a subjective score and used for the experiment.

We compare the performances of four algorithms: Hub, LID, Hub-first, and LID-first. The algorithm ``Hub'' ranks anchors based on their hub scores in descending order, and similarly for ``LID'' which performs ranking in ascending order of LID values. We assess the overall performance of top-K ranked anchors by averaging their subjective scores obtained in the user studies. The higher the average score is, the better an algorithm is. The result is presented in Table \ref{tab:1} using the combination of CNN, Concept and Text features. Overall, LID seems to be better coincident with human perception than hubness in evaluating the anchor qualities. Ranking anchors purely based on hub score runs into the risk of selecting noisy fragments, and the performance is consistently the worst across different values of top-K. 
\czq{Ranking anchors purely based on hub score runs into the risk of selecting noisy fragments. The performance is consistently the worst across different values of top-K, and sometimes is even worse than Random run.}
The proposed two algorithms, which takes into account the interplay among hubness, LID, and diversity, significantly outperforms the baselines Hub and LID. From the result, LID-first algorithm shows an edge over Hub-first. We believe that this is due to the bias in the user study, where fragments with one single theme are likely to receive higher subjective score than fragments with multi-perspective themes.
The result of LID-first is somewhat close to oracle, implying a certain degree of consistency between statistical analysis and human cognition of what should be an ideal anchor.

\begin{table}[!t]
	\centering
	\caption{
		\czq{Result of anchor selection. Oracle and Random are simulation runs, where the former shows the best possible performance and the latter selects anchor in random.}
	}
	\label{tab:1}
	\vspace{0.1in}
	\begin{tabularx}{8.5cm}{Xccc}
		\hline\hline
		\centering{}& \textbf{K=10}& \textbf{K=20} & \textbf{K=40}  \\
		\hline
		Hub          & 3.53 & 4.67 & 5.60 \\
		LID             & 5.93 & 6.13 & 6.60\\
		Hub-first           & 6.53   & 8.47 &  7.80\\
		LID-first             &7.40   & 8.26  &   8.33\\
		Oracle          &12.15   & 11.58  &   10.63\\
		Random    & 4.40 & 5.13 & 5.33\\
		\hline\hline
	\end{tabularx}
\end{table}


\begin{table}[!t]
	\centering
	\caption{\czq{Comparison of target selection with mAP}}
	\vspace{0.1in}
	\label{tab:2}
	\begin{tabularx}{8.5cm}{Xccc}
		\hline\hline
		\centering{}& \textbf{mAP@10}& \textbf{mAP@20}& \textbf{mAP@50}\\
		\hline
		Hub           &  0.06  & 0.09 &0.13 \\
		LID             & 0.08   & 0.11 &0.15 \\
		Hub-first     &  0.09 & 0.13  &0.17 \\
		LID-first   &  0.10  &0.13 &0.17  \\
		\hline\hline
	\end{tabularx}
\end{table}


\begin{table}[!t]
	\centering
	\caption{\wx{Performance of LID-first algorithm on different features. The \% in parenthesis indicates the relative improvement when LID-first is not in used. }}
	\vspace{0.1in}
	\label{tab:4}
	\begin{tabularx}{8.5cm}{llll}
		\hline\hline
		\centering{}& \textbf{mAP@10}& \textbf{mAP@20}& \textbf{mAP@50}\\
		\hline
		CNN        & 0.07 (1\%) & 0.11 (5\%)  & 0.13 (5\%)  \\ 
		Concept        &0.04 (216\%)  & 0.05 (163\%)   & 0.07 (106\%)  \\
		Text         &0.04 (46\%)  &0.05 (71\%)   & 0.08 (53\%)   \\
		CNN+Concept &0.08 (32\%)  &0.11 (26\%)   &0.14 (24\%) \\
		\hline
		LDA                  &0.02 (30\%)   &0.03 (42\%)   &0.05 (19\%)  \\
		CM\_LDA          &0.04 (-8\%)   &0.08 (19\%)  & 0.09 (18\% )  \\
		\hline
		CNN+\\Concept+Text   &0.10 (26\%)   &0.13 (23\%)  & 0.17 (28\%)   \\
		\hline\hline
	\end{tabularx}
\end{table}

\subsection{Target selection}
\czq{In this experiment, we treat each anchor as a query and conduct search of top-K target candidates.}
The top-k candidates are then re-ranked respectively by four different algorithms. As in TRECVid LNK, the performance is measured by mean average precision (mAP) at \czq{a} depth of $K$ in a rank list. 
\czq{The mAP is calculated over 28 anchors which have ground-truth targets provided by TRECVid. The number of ground-truth targets for each anchor ranges from 9 to 38. }
Table \ref{tab:2} shows the performances of four different algorithms based on the fusion of three features. Similar performance trend is observed as anchor selection.

We further show the effect of different features and their combinations for target selection. In addition to CNN, \czq{Concept} and \czq{Text}, the feature being considered are \czq{the topic-level} representation, which has been reported in \cite{15_CM_LDA} as reliable for video hyperlinking. Two kinds of topic modeling are considered: the classic Latent Dirichlet Allocation (LDA) \cite{03_LDA} using text features, and cross-modal LDA (CM-LDA) using Concept and CNN features based on the implementation of \cite{CM_LDA}. Table \ref{tab:4} lists the result of \czq{the} LID-first algorithm on different features. As can be seen, LID-first algorithm consistently introduce improvement for various features across different levels of depth. More importantly, such improvement is also noticed when different features are early fused, signifying the merit of manipulating hub, LID, and diversity for target selection.

\section{Discussion \& Conclusion}

We have presented the study of hubness and LID in quantifying the characteristics of anchors and targets. The study clearly shows the existence of hubness phenomenon in Blip10000 dataset. Particularly, the textual feature appears to be noisy resulting in fragments with extremely high hub scores. By fusion with deep and concept level features, the distribution of hubness is less skew implying a higher chance of achieving better performance with both text and visual features. Among the 122 anchors, 114 of them are regarded as hubs. Meanwhile, only \wx{4} of them have local intrinsic dimensionality higher than the global \wx{intrinsic} dimensionality of the entire dataset. When considering both hubness and LID, the majority of anchors are hubs with low to moderate risk in hyperlinking. By assuming that these are the required characteristics of anchors and targets, we take the bravery to propose the Hub-first and LID-first algorithms. Through the user study, surprisingly both algorithms somewhat align with human cognition of what could be think of as anchors. Furthermore, by applying the algorithms to re-rank targets, noticeable improvement is also attained across different modalities and their combinations. Basically, we can safely conclude that a fragment quantified as the hub with low local intrinsic dimensionality are likely to be a good anchor or target. Currently, it is still not clear of how the quality of anchors will correlate with the performance of target identification, which will be our future research direction.

\section{Acknowledgment}
This work was supported by the Research Grants Council of the Hong Kong Special Administrative Region, China (CityU 11250716), National Natural Science Foundation of China (61772436), Cultivation Program for the Excellent Doctoral Dissertation of Southwest Jiaotong University (2015310084), and Sichuan Science and Technology Innovation Seedling Fund (2017RZ0015 and 2017020).

{\small
	\bibliographystyle{ieee}
	\bibliography{egbib}
}

\end{document}